\documentclass[11pt]{article}
\usepackage{coling2020}
\usepackage{times}
\usepackage{url}
\usepackage{latexsym}
\usepackage{graphicx}
\usepackage{multibib} 
\usepackage{multirow}
\usepackage{amssymb}
\usepackage{pifont}
\usepackage{amsmath}
\usepackage{bm}

\colingfinalcopy 

\title{R-VGAE: Relational-variational Graph Autoencoder \\ for Unsupervised Prerequisite Chain Learning}

\author{Irene Li$^{1}$, Alexander Fabbri$^1$, Swapnil Hingmire$^2$ \and Dragomir Radev$^1$ \\
$^1$Yale University, USA \\
$^2$Tata Consultancy Services Limited (TCS), India
}

\date{}

\begin{document}
\maketitle
\begin{abstract}
 The task of concept prerequisite chain learning is to automatically determine the existence of prerequisite relationships among concept pairs. In this paper, we frame learning prerequisite relationships among concepts as an unsupervised task with no access to labeled concept pairs during training. We propose a model called the \textbf{R}elational-\textbf{v}ariational \textbf{G}raph \textbf{A}uto\textbf{E}ncoder (R-VGAE) to predict concept relations within a graph consisting of concept and resource nodes. Results show that our unsupervised approach outperforms graph-based semi-supervised methods and other baseline methods by up to 9.77\% and 10.47\% in terms of prerequisite relation prediction accuracy and F1 score. Our method is notably the first graph-based model that attempts to make use of deep learning representations for the task of unsupervised prerequisite learning. We also expand an existing corpus which totals $1,717$ English Natural Language Processing (NLP)-related lecture slide files and manual concept pair annotations over $322$ topics. 
\end{abstract}

\section{Introduction}

\begin{figure}[t]
\centering
\includegraphics[width=8.5cm]{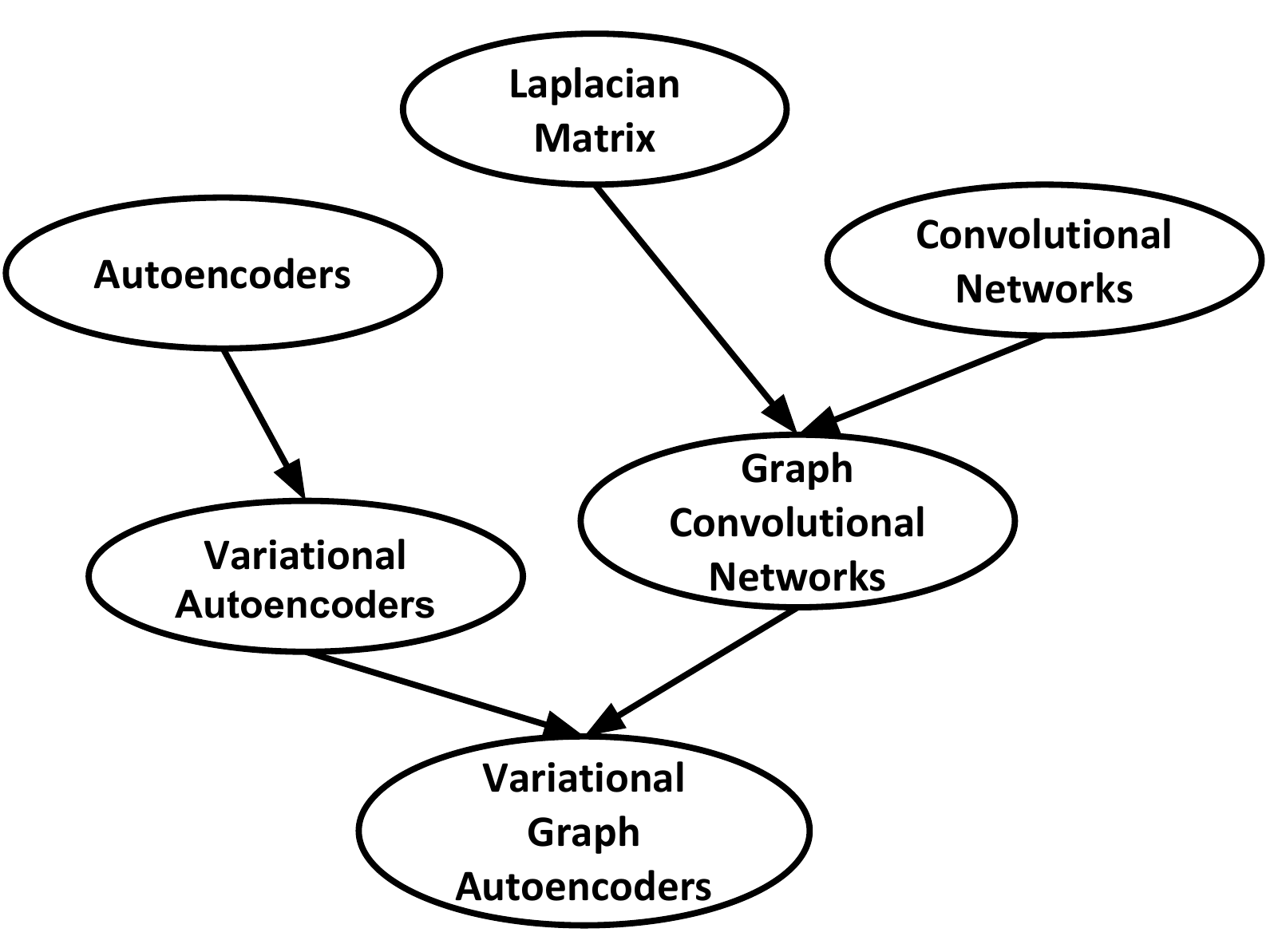}
\caption{An illustration of prerequisite chains: we show six concepts and the relations. For example, the concept \textit{Variational Autoencoders} is a prerequisite concept of  the concept \textit{Variational Graph Autoencoders}. }

\label{fig:intro}
\end{figure}

With the increasing amount of information available online, there is a rising need for structuring how one should process that information and learn knowledge efficiently in a reasonable order. As a result, recent work has tried to learn prerequisite relations among concepts, or which concept is needed to learn another concept within a \textit{concept graph}  \cite{liang2017recovering,gordon2016modeling,alsaad2018mining}. Figure \ref{fig:intro} shows an illustration of prerequisite chains as a directed graph. In such a graph, each node is a concept, and the direction of each edge indicates the prerequisite relation. Consider two concepts $p$ and $q$, we define $p\rightarrow q$ as $p$ is a prerequisite concept of $q$. For example, the concept \textit{Variational Autoencoders} is a prerequisite concept of  the concept \textit{Variational Graph Autoencoders}. If someone wants to learn about the concept \textit{Variational Graph Autoencoders}, the prerequisite concept \textit{Variation Autoencoder} should appear in the prerequisite concept graph in order to create a proper study plan. 

Recent work has attempted to extract such prerequisite relationships from various types of materials including Wikipedia articles, university course dependencies or MOOCs (Massive Open Online Courses) \cite{pan2017prerequisite,gordon2016modeling,liang2017recovering}. 
However, these materials either need additional steps for pre-processing and cleaning, or contain too many noisy free-texts, bringing more challenges to prerequisite relation learning or extracting. 
Recently, \newcite{li2018should} presented a collection of university lecture slide files mainly in NLP lectures with related prerequisite concept annotations. We expanded this dataset as we believe these lecture slides offer a concise yet comprehensive description of advanced topics.

Deep models such as word embeddings \cite{mikolov2013distributed} and more recently contextualized word embeddings \cite{devlin2018bert} have achieved great success in the NLP tasks as demonstrate a stronger ability to represent the semantics of the words than other traditional models. However, recent prerequisite learning approaches fail to make use of distributional semantics and advances in deep learning representations \cite{labutov2017semi,pan2017prerequisite}. In this paper, we investigate deep node embeddings within a graph structure to better capture the semantics of concepts and resources, in order to learn accurate the prerequisite relations.

In addition to learning node representations, there has been growing research in geometric deep learning \cite{bronstein17} and graph neural networks \cite{gori2005new}, which apply the representational power of neural networks to graph-structured data. Notably, \newcite{kipf2016semi} proposed Graph Convolutional Networks (GCNs) to perform deep learning on graphs, yielding competitive results in semi-supervised learning settings. TextGCN was proposed by \cite{yao2018graph} to model a corpus as a heterogeneous graph in order to jointly learn word and document embeddings for text classification. We build upon these ideas for constructing a resource-concept graph\footnote{We use the term \textit{resource} instead of \textit{document} for generalization.}.  Additionally, most of the mentioned methods require a subset of labels for training, a setting which is often infeasible in the real world. Limited research has been investigated learning prerequisite relations without using human annotated relations during training \cite{alsaad2018mining}. In practice, it is very challenging to obtain annotated concept-concept relations, as the complexity for annotating is $O(n^2)$ given $n$ concepts. 
To tackle this issue, we propose a method to learn prerequisite chains without any annotated concept-concept relations, which is more applicable in the real word.

Our contributions are two-fold: 1) we expand upon previous annotations to increase coverage for prerequisite chain learning in five categories, including AI (artificial intelligence), ML (machine learning), NLP, DL (deep learning) and IR (information retrieval). We also expand a previous corpus of lecture files to include an additional 5000 more lecture slides, totaling 1,717 files. More importantly, we add additional concepts, totaling 322 concepts, as well as the corresponding annotations of each concept pair, which totals 103,362 relations.  2) we present a novel graph neural model for learning prerequisite relations in an unsupervised way using deep representations as input. 
We model all concepts and resources in the corpus as nodes in a single heterogeneous graph and define a propagation rule to consider multiple edge types by eliminating concept-concept relations during training, making it possible to perform unsupervised learning.
Our model leads to improved performance over a number of baseline models. 
Notably, it is the first graph-based model that attempts to make use of deep learning representations for the task of unsupervised prerequisite learning. Resources, annotations and code are publicly available online\footnote{\url{https://github.com/Yale-LILY/LectureBank/tree/master/LectureBank2}}.

\section{Related Work}

\subsection{Deep Models for Graph-structured Data}

There has been much research focused on graph-structured data such as social networks and citation networks \cite{sen2008collective,akoglu2015graph,defferrard2016convolutional}, and many deep models have achieved satisfying results. Deepwalk \cite{deepwalk} was a breakthrough model which learns node representations using random walks. Node2vec \cite{node2vec-kdd2016} was an improved scalable framework, achieving promising results on multi-label classification and link prediction. Besides, there has been some work like graph convolution neural networks (GCNs), which target on deep-based propagation rules within graphs. 
A recent work applied GCN for text classification \cite{yao2018graph} by constructing a single text graph for a corpus based on word co-occurrence and document word relations. The experimental results showed that the proposed model achieved state-of-the-art methods on many benchmark datasets. We are inspired by this work in that we also attempt to construct a single graph for a corpus, however, we have different types of nodes and edges. 



\subsection{Prerequisite Chain Learning}

Learning prerequisite relations between concepts has attracted much recent work in machine learning and NLP field. Existing research focuses on machine learning methods (i.e., classifiers) to measure the prerequisite relations among concepts \cite{liang2018investigating,liu2016learning,liang2017recovering}. Some research integrates feature engineering to represent a concept, inputting these features to a classic classifier to predict relationship of a given concept pair \cite{liang2017recovering,liang2018investigating}. The resources to learn those concept features include university course descriptions and materials as well as online educational data \cite{liu2016learning,liang2017recovering}. Recently, \newcite{li2018should} introduced a dataset containing 1,352 English lecture files collected from university-level lectures as well as 208 manually-labeled prerequisite relation topics, initially introduced in \cite{fabbri2018tutorialbank}. To avoid feature engineering, they applied graph-based methods including GAE and VGAE \cite{kipf2016semi} which treat each concept as a node thus building a concept graph. They pretrained a Doc2vec model \cite{le2014distributed} to infer each concept as a dense vector, and then trained the concept graph in a semi-supervised way. Finally, the model was able to recover unseen edges of a concept graph. Different from their work, we wish to do the prerequisite chain learning in an unsupervised manner, while in training, no concept relations will be provided to the model.

{
\setlength{\abovedisplayskip}{0pt}
\setlength{\belowdisplayskip}{0pt}
\begin{table}[t] 
    \centering
\begin{tabular}{ l c c c c c }
\hline \hline
\textbf{Domain}    & \textbf{\#courses} & \textbf{\#files} & \textbf{\#tokens}   & \textbf{\#pages} & \textbf{\#tokens/page} \\ \hline 
NLP     & 45       & 953    & 1,521,505  & 37,213 & 40.886    \\ 
ML      & 15       & 312    & 722,438   & 12,556 & 57.537 \\
DL      & 7        & 259    & 450,879    & 7,420  & 60.765  \\ 
AI      & 5        & 98     & 139,778    & 3,732  & 37.454   \\ 
IR      & 5        & 95     & 205,359  & 4,107  & 50.002 \\ \hline
\textbf{Overall} & 77       & 1,717  & 3,039,959  & 65,028 & 46.748   \\	\hline \hline
\end{tabular}
    \caption{Dataset Statistics. 
    In each category, we have a given number of courses (\#courses); each course consists of lecture files (\#files); each lecture file has a number of individual slides (\#pages). We also show the number of total tokens (\#tokens) and average token number per slide (\#tokens/page). }
    \label{tab:courses}
\end{table}

}

\section{Dataset}

\subsection{Resources}
We manually collected English lecture slides mainly on NLP-related courses in recent years from known universities. We treated them as individual slide file in PDF or PowerPoint Presentations format. Our new collection has 529 additional files from 17 courses, which we combined with the data provided by \cite{li2018should}. We ended up with a total number of 77 courses with 1,717 English lecture slide files, covering five domains. We show the final statistics in Table \ref{tab:courses}. For our experiments, we converted those files into TXT format which allowed us to load the free texts directly.

\subsection{Concepts}
We manually expanded the size of concept list proposed by \cite{li2018should} from 208 to 322. We included concepts which were not found in their version like \textit{restricted boltzmann machine} and \textit{neural parsing}. Also, we re-visited their topic list and corrected a small number of the topics. For example, we combined certain topics (e.g. \textit{BLUE} and \textit{ROUGE}) into a single topic (\textit{machine translation evaluation}).  
We asked two NLP PhD students to re-evaluate existing annotations from the old corpus and to provide labels for each added concept pair in the new corpus. A Cohen kappa score \cite{cohen1960coefficient} of 0.6283 achieved between our annotators which can be considered as a substantial agreement. We then took the union of the annotations, where if at least one judge stated that a given concept pair $(A,B)$ had $A$ as a prerequisite of $B$, then we define it a positive relation. We believe that the union of annotations makes more sense for our downstream application, where we want users to be able to mark which concepts they already know and displaying all potential concepts is essential. We have 1,551 positive relations on the 322 concept nodes.


\section{Method}

\begin{figure*}[t]
\centering
\small
\includegraphics[width=12cm]{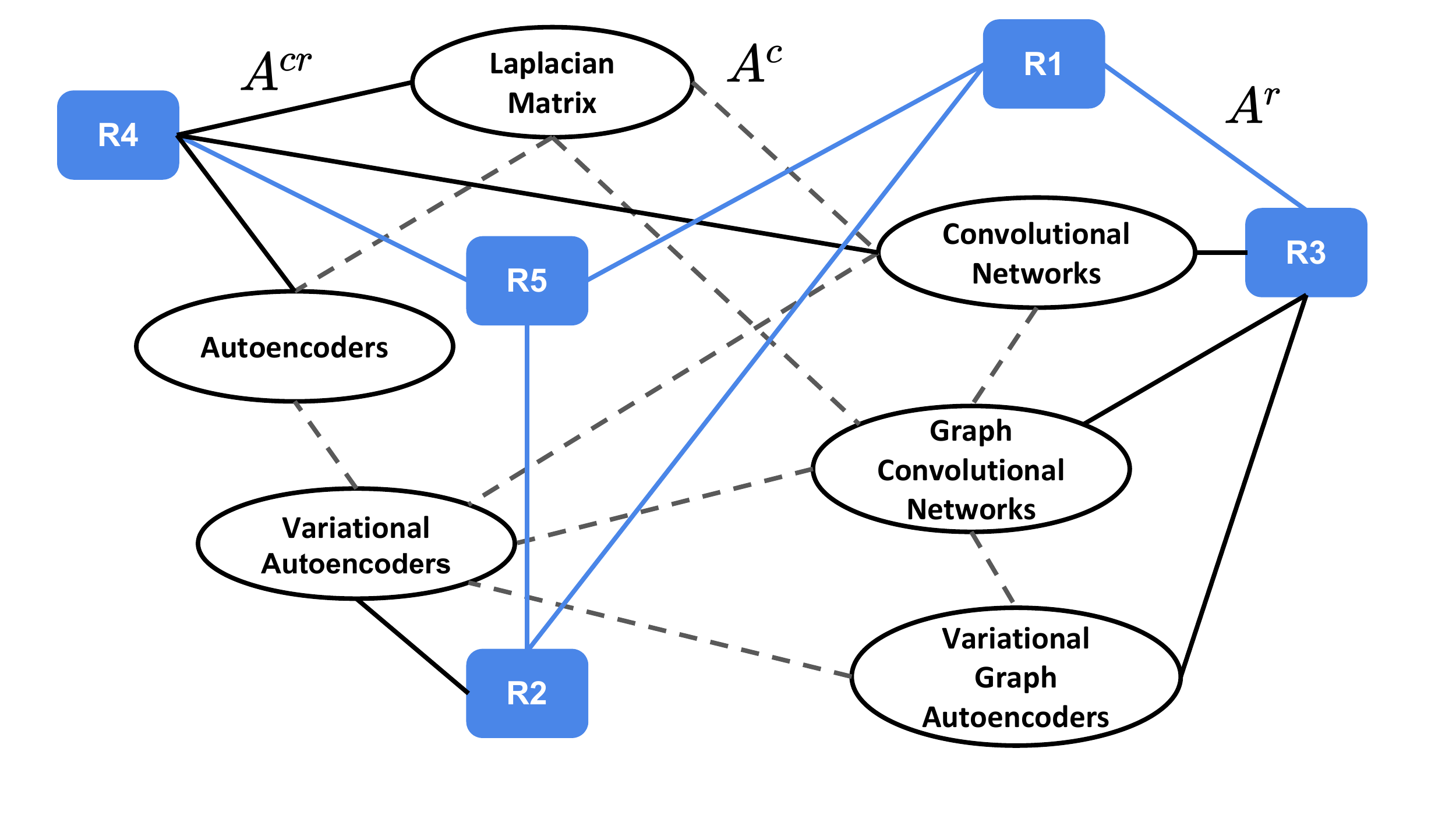}

\caption{Concept-resource graph for prerequisite chain learning: oval nodes indicate concept nodes, the blue rectangular nodes indicate resource nodes. We show three types of edges: the blue edge between two resource nodes $A^r$, the black solid edge between a concept node and a resource node $A^{cr}$ and the black dashed edge between two concept nodes $A^c$. In the graph, the resource nodes R1 to R5 are example resources used to illustrate the idea. In practice there may be more edges, we show a part of the them for simplicity. }

\label{fig:model}
\end{figure*}

\subsection{Problem Definition}

In our corpus, every concept $c$ is a single word or a phrase; every resource $r$ is free text extracted from the lecture files. We then wish to determine for a given concept pair $(c_i,c_j)$, whether $c_i$ is a prerequisite concept of $c_j$. 
We define the concept-resource graph as $G=(X, A)$, where $X$ denotes node features or representations and $A$ denotes the adjacency matrix. In our case, the adjacency matrix is the set of relations between each node pair, or the edges between the nodes.
In Figure~\ref{fig:model}, we build a single, large graph consisting of concepts (oval nodes) and resources (rectangular nodes) as nodes, and the corresponding relations as edges. So there are three types of edges in $A$: the edge between two resource nodes $A^r$ (blue line), the edge between a concept node and a resource node $A^{cr}$ (black solid line), and the edge between two concept nodes $A^c$ (black dashed line). 
Our goal is to learn the relations between concepts only ($A^c$), so prerequisite chain learning can be formulated as a link prediction problem. Our unsupervised setting is to exclude any direct concept relations ($A^c$) during training, and we wish to predict these edges through message passing via the resource nodes indirectly. 


\subsection{Preliminaries}

\textbf{Graph Convolutional Networks} (GCN) \cite{kipf2016semi} is a semi-supervised learning approach for node classification on graphs. It aims to learn the node representation $H=\{h_1,h_2,..h_n\}$ in the hidden layers, given the initial node representation $X$ and the adjacency matrix $A$. The model incorporates local graph neighborhoods to represent a current node. In a simple GCN model, a layer-wise propagation rule can be defined as the following:

\begin{equation}
    H^{(l+1)} = \sigma(AH^{(l)}W^{(l)})
\label{eq:gcn1}
\end{equation}

where $l$ is the current layer number, $\sigma(\cdot)$ is a  non-linear activation function, and $W$ is a parameter matrix that can be learned during training. We eliminate the $\sigma(\cdot)$ for the last layer output. For the task of node classification, the loss function is cross-entropy loss. Typically, a two-layer GCN (by plugging Equation \ref{eq:gcn1} in) is defined as:

\begin{equation}
   GCN(X,A) = H^2= \tilde{A}H^1W^1=\tilde{A}\sigma(AXW^0)W^1
   \label{eq:gcn2}
\end{equation}

where $\tilde{A}$ is the new adjacency matrix at the second graph layer. 

\textbf{Relational Graph Convolutional Networks} (R-GCNs) \cite{schlichtkrull2018modeling} expands the types of graph nodes and edges based on the GCN model, allowing operations on large-scale relational data. In this model, an edge between a node pair $i$ and $j$ is denoted as $(v_i,rel, v_j)$, where $rel \in Rel$ is considered a relation type, while in GCN, there is only one type. Similarly, to obtain the hidden representation of the node $i$, we consider the local neighbors and itself; when multiple types of edges exist, different sets of weight will be considered. So the layer-wise propagation rule is defined as:



{
\setlength{\abovedisplayskip}{0pt}
\setlength{\belowdisplayskip}{0pt}
\begin{equation}
\small
h_{ i }^{ (l+1) }=\sigma \left(\frac{1}{M}\sum _{ rel \in Rel} \sum _{ j\in { N }_{ i }^{ rel } } (W_{ r }^{ (l) }h_{ j }^{ (l) }+W_{ 0 }^{ (l) }h_{ i }^{ (l) })  \right) 
\label{eq:rgcn}
\end{equation}
}
where $Rel$ is the set of relations or edge types in the graph, $N_{i}^{rel}$ denotes the neighbors of node $i$ with relation $rel$, $W_r^{(l)}$ is the weight matrix at layer $l$ for nodes in $N_{i}^{rel}$, $W_0^{(l)}$ is the shared weight matrix at layer $l$, $M$ is the number of weight matrices in each layer.



\textbf{Variational Graph Auto-Encoders} (V-GAE) \cite{kipf2016variational} is a framework for unsupervised learning on graph-structured data based on variational auto-encoders \cite{kingma2013auto}. 
It takes the adjacency matrix and node features as input and tries to recover the graph adjacency matrix $A$ through the hidden layer embeddings $Z$. Specifically, the non-probabilistic graph auto-encoder (GAE) model calculates embeddings via a two-layer GCN encoder: $Z=GCN(X,A)$, which is given by Equation \ref{eq:gcn2}. 

Then, in the variational graph auto-encoder, the goal is to sample the latent parameters $z_i\in Z$ from a normal distribution: 
\begin{equation}
\begin{aligned}
q\left(z_{i} | X, A\right) &=\mathcal{N}\left(z_{i} | \boldsymbol{\mu}_{i}, \operatorname{diag}\left(\sigma_{i}^{2}\right)\right)
\label{eq:inference2}
\end{aligned}
\end{equation}

where $\mu=\mathrm{GCN}_{\mu}(X, A)$ is the matrix of mean vectors, and $\log_{\sigma}=\mathrm{GCN}_{\sigma}(X, A)$. The training loss then is given as the KL-divergence between the normal distribution and the sampled parameters $Z$:

\begin{equation}
    \mathcal{L}_{\text {latent }}=\sum_{i \in \mathcal{N}} \operatorname{KL}\left(\mathcal{N}\left(\boldsymbol{\mu}_{i}, \operatorname{diag}\left(\boldsymbol{\sigma}_{i}\right)^{2}\right) \| \mathcal{N}(\mathbf{0}, \mathbf{I})\right)
    \label{eq:latent}
\end{equation}

In the inference stage, the reconstructed adjacency matrix $\hat{A}$ is the inner product of the latent parameters $Z$: $\hat{A}=\sigma(ZZ^T)$.


\subsection{Proposed Model}
To take multiple relations into consideration and make it possible to do unsupervised learning for concept relations, we propose our R-VGAE model. Our model builds upon R-GCN and VGAE by taking the advantages of both: R-GCN is a supervised model that deals with multiple relations; VGAE is an unsupervised graph neural network. We then make it possible to directly to train on a heterogeneous graph in an unsupervised way for link prediction, in order to learn the prerequisite relations for the concept pairs. 

Our model first applies the R-GCN in Equation \ref{eq:rgcn} as the encoder to obtain the latent parameters $Z$, given the initial node features $X$ and adjacency matrix $A$: $Z=\textrm{R-GCN}(X,A)$. In terms of the variational verison, as opposed to the standard VGAEs, we parameterize $\mu$ by the RGCN model: $\mu=\textrm{R-GCN}_{\mu}(X, A)$, and $\log_{\sigma}=\textrm{R-GCN}_{\sigma}(X, A)$.


To predict the link between a concept pair $(c_i,c_j)$, we followed the DistMult \cite{yang2014embedding} method: we take the last layer output node features $\hat{X}$, and define the following score function to recover the adjacency matrix $\hat{A}$ by learning a trainable weight matrix $R$:

\begin{equation}
\hat{A}= \hat{X}^{\intercal} R \hat{X}
\label{eq:link}
\end{equation}

The loss consists of the cross-entropy reconstruction loss of adjacency matrix ($\mathcal{L}_{cross}$) and the loss from the latent parameters defined in Equation \ref{eq:latent}:

\begin{equation}
\mathcal{L} = \mathcal{L}_{cross}(A,\hat{A}) + \mathcal{L}_{latent}
\label{eq:loss}
\end{equation}


We compare two variations of our R-GAE model. \textbf{Unsupervised}: only the concept-resource edges $A^{cr}$ and resource-resource edges $A^{r}$ are provided during training. This is an \textit{unsupervised} model because no concept-concept edges are used. \textbf{Semi-supervised}: the model has access to  concept-resource edges $A^{cr}$ and resource-resource edges $A^{r}$, as well as a percentage of the available concept-concept edges $A^{c}$, described later.

\subsection{Node Features $X$}
\textbf{Sparse Embeddings}
We used TFIDF (term frequency–inverse document frequency) to get sparse embeddings for all nodes. We restricted the global vocabulary to be the 322 concept terms only, which means that the dimension of the node features is 322, as we aim to model keywords. 

\textbf{Dense Embeddings} \label{dense}
As the concepts in our corpus often consist of phrases such as \textit{dynamic programming}, we made use of Phrase2vec \cite{artetxe2018unsupervised}. Phrase2vec (P2V) is a generalization of skip-gram models \cite{mikolov2013distributed} which learns n-gram embeddings during training, and here we aim to infer the embeddings of the concepts in our corpus. We trained the P2V model using only our corpus by treating each slide file as a short document as a sequence of tokens. For each resource node, we take an element-wise average of the P2V embeddings of each single token and phrases that resource covered. Similarly, for each concept node, we took element-wise average of the embeddings of each individual token and the concept phrase. 
In addition, we then utilized the BERT model \cite{devlin2018bert} as another type of dense embedding. We fine-tuned the masked language modeling of BERT using our corpus. 


\subsection{Adjacency Matrix $A$}
To construct the adjacency matrix $A$, for each node pair $(v_i,v_j)$, we applied cosine similarity based on enriched TFIDF features\footnote{This means that the TFIDF features are calculated on an extended vocabulary that includes all possible tokens appeared in the corpus.} as the value $A_{ij}$. Previous work has applied cosine similarity for vector space models \cite{garcia2018w2vlda,zuin2018learning,bhatia2016automatic}, so we believe it is a suitable method in our case. This way we were able to generate concept-resource edge values ($A^{cr}$) and resource-resource edge values ($A^{r}$). Note that for concept-concept edge values $A^{c}$: 1 if $c_i$ is a prerequisite of $c_j$, 0 otherwise. These values are not computed in the unsupervised setting.



\section{Evaluation}

We compare our proposed model with two groups of baseline models. We report accuracy, F1 scores, the macro averaged Mean Average Precision (MAP) and Area under the ROC Curve (AUC) scores in Table \ref{tab:res_10}, as done by previous research \cite{chaplot2016data,pan2017prerequisite,li2018should}. We split the $1,551$ positive relations into 9:1 (train/test), and randomly select negative relations as negative training samples, and then we run over five random seeds and report the average scores, following the same setting with \newcite{kipf2016variational} and \newcite{li2018should}. 

\begin{table}[t] 
    \centering
\begin{tabular}{lcccc} \hline \hline
     \textbf{Method}        & \textbf{Acc}     & \textbf{F1}      & \textbf{MAP}     & \textbf{AUC}     \\   \hline
\multicolumn{5}{l}{\textbf{\textit{Concept embedding + classifier}} }\\
P2V (lb1) &	0.5927 &	0.5650 &	0.5623 &	0.5929 \\
P2V (lb2) &	\underline{0.6369} &	\underline{0.5961} &	\underline{0.6282} &	\underline{0.6370} \\
BERT (lb1) & 0.6540    &	0.6099 &	0.6475 &	0.6540 \\
BERT (lb2) & \underline{0.6558} &	\underline{0.6032} &	\underline{0.6553} &	\underline{0.6558} \\
BERT (original) &	\textbf{0.7088} &	\textbf{0.6963} &	\textbf{0.6779} &	\textbf{0.7090}    
       \\ \hline 
\multicolumn{5}{l}{\textbf{\textit{Graph-based methods}}} \\
DeepWalk \cite{deepwalk}   & 0.6292 & 0.5860	& 0.6270 &	0.6281 \\
Node2vec \cite{node2vec-kdd2016} & 0.6209 & 0.6181	& 0.5757 &	0.6259\\
VGAE \cite{li2018should}       & 0.6055          & 0.6030         & 0.5628         & 0.6055         \\
GAE \cite{li2018should}& \textbf{0.6307} & \textbf{0.6275} & \textbf{0.5797} & \textbf{0.6307} \\ 
R-GCN \cite{schlichtkrull2018modeling}	& 0.5387	& 0.4784	& 0.5203	& 0.5387 \\
\hline
\multicolumn{5}{l}{\textbf{\textit{R-VGAE (Our proposed model)}}} \\
US+BERT (fine-tuned) & 0.5704 &	0.5704 &	0.5579 &	0.5955 \\
US+BERT (original)  & 0.5669 &	0.5668 &	0.5658 &	0.6164 \\
US+TFIDF & 0.6495          &0.6458	 &  0.7069	& 0.5507          \\
US+P2V   & 0.7694*          & 0.7638*          & 0.8919*          & 0.9126*          \\
SS+BERT (fine-tuned) & 0.6942 & 	0.6942 &	0.6613 &	0.7412 \\
SS+BERT (original) & 0.6839 &	0.6839 &	0.6556 &	0.7372 \\
SS+TFIDF   & 0.7252          & 0.7082          & 0.8181          & 0.7625          \\
SS+P2V      & \textbf{0.8065} & \textbf{0.8010} & \textbf{0.9380} & \textbf{0.9454} \\ \hline \hline
\end{tabular}
\caption{Accuracy (\textit{Acc}), macro F1, MAP and AUC scores on balanced test set including 10\% of prerequisite edges. Bold values are the best results within its experiment group. Underscored values indicate a better performance compared with the trained corpus. Values with an asterisk mean the best performance in the unsupervised setting.}
\label{tab:res_10}
\end{table}

\textbf{Concept embedding + classifier} The first group is the concept embedding with traditional classifiers including Support Vector Machines, Logistic Regression, Na\"ive Bayes and Random Forest. For a given concept pair, we concatenate the dense embeddings for both concepts as input to train the classifiers, and then we report the best result. We compare Phrase2Vec (P2V) and BERT embeddings. We have two corpora: one is the old version (\textit{lb1}) provided by \cite{li2018should}, another one is our version (\textit{lb2}). For the BERT model, we applied both the original version from Google (\textit{original}) \footnote{\url{https://github.com/google-research/bert}}, and the fine-tuned language models version on our corpora (\textit{lb1, lb2}) from \newcite{xiao2018bertservice}, and perform inference on the concepts. The P2V embeddings have 150 dimension, and the BERT embeddings have 768 dimensions. We show improvements on the BERT and P2V baselines by using our additional data via the underscored values. This indicates that the concept relations can be more accurately predicted when enriching the training corpus to train better embeddings. In our following experiments, if not specified, we applied \textit{lb2} as the training corpus. 

\textbf{Graph-based methods} 
We apply the classic graph-based embedding methods DeepWalk \cite{deepwalk} and Node2vec \cite{node2vec-kdd2016}, by considering the concept nodes only. Then the positive concept relations in training set are the known sequences, allowing to train both models to infer node features. Similarly, in the testing phrase, we concatenate the node embeddings given a concept pair, and utilize the mentioned classifiers to predict the relation and report the performance of the best one. 
We then include VGAE and GAE methods for prerequisite chain learning following \newcite{li2018should}. Both methods construct the concept graph in a semi-supervised way. We apply P2V embeddings to replicate their methods, though it is possible to try additional embeddings, this is not our main focus. Finally, we compare with the original R-GCN model for link prediction proposed by \newcite{schlichtkrull2018modeling} and apply the same embeddings with the VGAE and GAE methods. 
Other semi-supervised graph methods such as GCNs require node labels and thus are not applicable to our setting. We can see that the GAE method achieves the best results among the baselines. Compare with the first group, BERT (\textit{original}) still has a better performance due to its ability to represent phrases.

\textbf{R-VGAE} Our model can be trained in both unsupervised (\textit{US+*}) and semi-supervised (\textit{SS+*}) way. We also utilize various types of embeddings include P2V, TFIDF, BERT (fine-tuned) and BERT (original). The best performed model in the unsupervised setting is with P2V embeddings, marked with asterisks, and it is better than all the baseline supervised methods with a large margin. In addition, our semi-supervised setting models boost the overall performance. We show that the \textit{SS+P2V} model performs the best among all the mentioned methods, with a significant improvement of 9.77\% in accuracy and 10.47\% in F1 score compared with the best baseline model \textit{BERT (original)}. This indicates that R-VGAE model does better on link prediction by bringing extra resource nodes into the graph, while the concept relation can be improved and enhanced indirectly via the connected resource nodes. 
We also observe that with BERT embeddings, the performance lags behind the other embedding methods for our approach. A reason might be that the dimensionality of the BERT embeddings is relatively large compared to P2V and may cause overfitting, especially when the edges are sparse; and it might not suitable to represent resources as they are a list of keywords when fine-tuning the language modeling. The P2V embeddings outperform TFIDF for both unsupervised and semi-supervised models. This shows that compared with sparse embeddings, dense embeddings can better preserve the semantic features when integrated within the R-GAE model, thus boosting the performance. Besides, as a variation of R-GCN and GAE, our model surpasses them by taking the advantages of both, comparing with R-GCN and GAE results reported in the second group.

\section{Analysis}

\begin{table*}[t!] 
    \centering
\begin{tabular}{c|c|c} \hline \hline
          \textbf{Concept}    & \textbf{Gold Prerequisite Concepts}     & \textbf{Model Output Concept}    \\ \hline
 
\multirow{8}{0.11\textwidth}{\textit{dependency parsing}} & syntax	&	syntax	\\
&classic parsing methods	&	classic parsing methods	\\
&linguistics basics	&	linguistics basics	\\
&parsing	&	parsing	\\
&nlp introduction	&	nlp introduction	\\
&chomsky hierarchy	&	chomsky hierarchy	\\
&linear algebra	&	linear algebra	\\
&conditional probability	&		\\

\hline 
 \multirow{8}{0.11\textwidth}{\textit{tree adjoining grammar}} &classic parsing methods	&	classic parsing methods	\\
&linguistics basics	&	linguistics basics	\\
&parsing	&	parsing	\\
&nlp introduction	&	nlp introduction	\\
&context free grammar	&	context free grammar	\\
&probabilistic context free grammars	&	probabilistic context free grammars	\\
&chomsky hierarchy	&	chomsky hierarchy	\\
&context sensitive grammar	&	context sensitive grammar	\\ \hline \hline

\end{tabular}
\small
\caption{A comparison of prerequisite concepts of 
\textit{dependency parsing} (upper group) and \textit{tree adjoining grammar} (lower group) from our annotated gold labels and labels recovered by our best unsupervised model.}

\label{tab:ana}
\end{table*}


We then take the recovered concept relations from our best performed model \textit{R-VGAE (SS+P2V)} in Table \ref{tab:res_10}), and compare them with the gold annotated relations. Note that here we only look at concept nodes. The average degree for gold graph concept nodes is 9.79, while our recovered one has an average degree of 6.10, and this means our model predicts fewer edges. We also check the most popular concepts that have the most degrees. We select \textit{dependency parsing} and \textit{tree adjoining grammar} as examples. In Table \ref{tab:ana}, we show a comparison of the prerequisites from the annotations and our model's output. The upper group illustrates results for \textit{dependency parsing}, where one can notice that the predicted concepts all appear in the gold results, missing only a single concept. This shows that even though our model predicts less number of relations, it still predicts correct relations. The lower group shows the comparison for the concept \textit{tree adjoining grammar}, our model gives precise prerequisite concepts among all eight concepts from the gold set. When a concept has a certain amount number of prerequisite concepts, our model is able to provide a comprehensive concept set with a good quality. In the real word, especially in a learner's scenario, he or she wants to learn the new concept with enough prerequisite knowledge, which our model tends to provide.

\section{Conclusion and Future Work}

In this paper we introduced an expanded dataset for prerequisite chain learning with additional an 5,000 lecture slides, totaling 1,717 files. We also provided prerequisite relation annotations for each concept pair among 322 concepts. Additionally, we proposed an unsupervised learning method which makes use of advances in graph-based deep learning algorithms. Our method avoids any feature engineering to learn concept representations. Experimental results demonstrate that our model performs well in an unsupervised setting and is able to further benefit when labeled data is available. In future work, we would like to perform a more comprehensive model comparison and evaluation by bringing other possible variations of graph-based models to learn a concept graph. Another interesting direction is to apply multi-task learning to the proposed model by adding a node classification task if there are node labels available. A part of the future work would also include developing educational applications for learners to find out their study path for certain concepts.


\bibliographystyle{coling}
\bibliography{coling2020}

\clearpage

\appendix

\end{document}